\documentclass{article}
\usepackage[utf8]{inputenc}
\usepackage{graphicx}
\usepackage{amsmath}
\usepackage{booktabs}
\usepackage{multirow}
\usepackage[hidelinks]{hyperref}
\usepackage{geometry}
\usepackage{caption}
\usepackage{subcaption}
\usepackage{float}
\usepackage[numbers,sort&compress]{natbib} 
\usepackage{url} 

\title{Selecting Optimal Camera Views for Gait Analysis: A Multi-Metric Assessment of 2D Projections}

\author{
Dong Chen\textsuperscript{1,2} \and
Huili Peng\textsuperscript{1} \and
Yong Hu\textsuperscript{1,2} \and
Kenneth MC. Cheung\textsuperscript{1,2,\thanks{Correspondence: cheungmc@hku.hk}}
}

\begin{document}

\maketitle
\thispagestyle{empty} 

\date{}

\thanks{%
\textsuperscript{1}Orthopaedic Centre, The University of Hong Kong Shenzhen Hospital, Shenzhen, China. \\
\textsuperscript{2}Department of Orthopaedics \& Traumatology, Li Ka Shing Faculty of Medicine, The University of Hong Kong, Hong Kong, China. 
}

\begin{abstract}
\noindent \textbf{Objective:} To systematically quantify the effect of the camera view (frontal vs. lateral) on the accuracy of 2D markerless gait analysis relative to 3D motion capture ground truth. \\
\textbf{Methods:} Gait data from 18 subjects were recorded simultaneously using frontal, lateral and 3D motion capture systems. Pose estimation used YOLOv8. Four metrics were assessed to evaluate agreement: Dynamic Time Warping (DTW) for temporal alignment, Maximum Cross-Correlation (MCC) for signal similarity, Kullback-Leibler Divergence (KLD) for distribution differences, and Information Entropy (IE) for complexity. Wilcoxon signed-rank tests (significance: $p < 0.05$) and Cliff's delta ($\delta$) were used to measure statistical differences and effect sizes. \\
\textbf{Results:} Lateral views significantly outperformed frontal views for sagittal plane kinematics: step length (DTW: $53.08 \pm 24.50$ vs. $69.87 \pm 25.36$, $p = 0.005$) and knee rotation (DTW: $106.46 \pm 38.57$ vs. $155.41 \pm 41.77$, $p = 0.004$). Frontal views were superior for symmetry parameters: trunk rotation (KLD: $0.09 \pm 0.06$ vs. $0.30 \pm 0.19$, $p < 0.001$) and wrist-to-hipmid distance (MCC: $105.77 \pm 29.72$ vs. $75.20 \pm 20.38$, $p = 0.003$). Effect sizes were medium-to-large ($\delta: 0.34$--$0.76$). \\
\textbf{Conclusion:} Camera view critically impacts gait parameter accuracy. Lateral views are optimal for sagittal kinematics; frontal views excel for trunk symmetry. \\
\textbf{Significance:} This first systematic evidence enables data-driven camera deployment in 2D gait analysis, enhancing clinical utility. Future implementations should leverage both views via disease-oriented setups.
\end{abstract}


\section{Introduction}
GAIT analysis is an essential tool in healthcare and medicine, enabling clinicians to identify disease-specific gait abnormalities, optimise treatment strategies, and monitor patient outcomes \cite{Gagnat2022, Khalid2023, Okuma2014, DiBiase2020, Cicirelli2022, Brognara2021, Pirker2017, Hu2024,Tangjade2024}. Recent advancements in marker-based 3D motion capture technologies and markerless-based 2D pose estimation systems have significantly improved the accessibility, accuracy, and cost-efficiency of gait analysis. Traditional marker-based systems, such as Vicon or Microsoft Kinect \cite{Nairn2025, Choi2024, Stenum2021, Wade2022, Nakano2020, Graff2024, Matsuda2025, Bernal2024, Yazdi2024}, and wearable devices  \cite{ Cicirelli2022, Brognara2021, Chen2011, Bowman2024, Cani2025} require extensive pre-processing, precise sensor placement and controlled environments, which often lead to increased complexity, higher costs, and limited applicability in real-world settings.

Recently, markerless-based systems have emerged as a critical tool in clinical gait analysis \cite{Carvalho2024, PantzarCastilla2024, DHaene2024, Anderson2025}, enabling healthcare professionals to quantitatively assess and better understand patients' gait patterns \cite{Stenum2021b}. Video-based pose estimation systems facilitate quantitative gait analysis from multiple views, including frontal and lateral planes \cite{Dubey2023, Eguchi2024, Mundt2024}. One study highlights deep learning-based pose estimation techniques' superior performance across various domains \cite{Zheng2023}. Specifically, these 2D pose estimation technologies offer practical solutions for real-world applications. Certain studies have validated their accuracy in measuring key kinematic gait factors such as gait cycle time, step length, walking speed, and sagittal plane joint angles \cite{Cao2017, Celik2024, Stenum2024}. Research highlighted their effectiveness, with lateral views excelling in kinematic analysis and frontal views aiding gait recognition using RGB-D cameras \cite{Vafadar2022, Chattopadhyay2014}. These approaches have offered cost-effective solutions for real-world clinical applications without the need for expensive equipment and controlled laboratory environments.

\begin{figure}[H]
\centering
\includegraphics[width=0.9\textwidth]{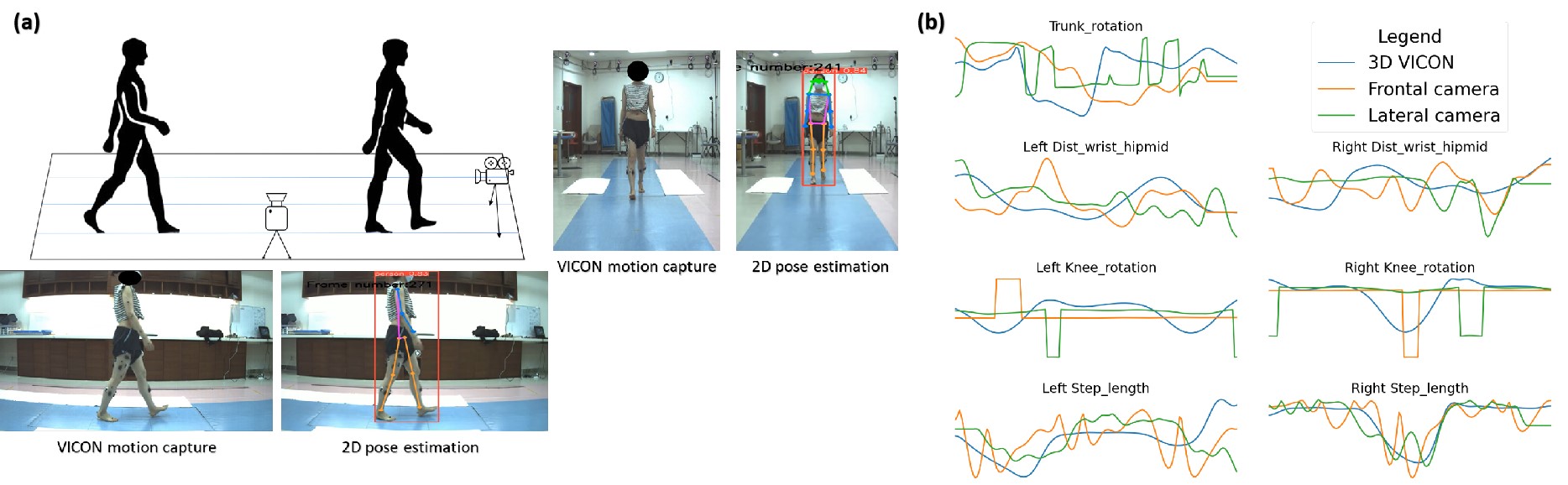} 
\caption{(a) The comparison of gait data from different perspectives, including 3D motion capture (VICON) and 2D projections from frontal and lateral views. (b) The figure highlights the variability in trunk rotation, wrist-to-midpoint of hipline, and knee rotation parameters across the three configurations. Other signal graphs are in the supplementary materials.}
\label{fig:fig1}
\end{figure}

However, despite advancements in 2D markerless systems, interpreting 3D gait dynamics from these methods remains challenging due to the inherent loss of spatial information during the projection of 3D motion into 2D views, as illustrated in  Fig.~\ref{fig:fig1}. Whereas the general trends of the 2D and 3D manifold data appear similar, the fine-grained patterns exhibit noticeable discrepancies. These differences raise critical questions about the clinical reliability of gait parameters derived from different camera views. 

The impact of video recording views on gait evaluation has been a key topic in the literature \cite{Pattanapisont2024, Cheng2025}. Previous studies have introduced workflows for analysing videos from various views, such as frontal and sagittal planes, and demonstrated the robustness of tools in estimating gait parameters across viewpoints \cite{Nakano2020, Stenum2024}. These findings highlight the critical role of camera angle selection in achieving accurate and reliable gait analysis. Despite these advances, study lacked a detailed exploration of the 2D data patterns derived from different views and do not adequately capture the intrinsic temporal, spatial, and kinematic complexities of human gait  \cite{Stenum2024}. Additionally, current metrics for evaluating 2D data fail to address key factors such as variability, scale, orientation, occlusion, and temporal coherence across views. This gap limits the ability to fully model multi-view interdependencies and hinders robust comparisons between data captured from frontal and lateral views. 

Whereas prior studies have qualitatively acknowledged view-dependent differences, none have systematically quantified these effects across temporal, statistical, and information-theoretic dimensions \cite{Stenum2021, Stenum2024, Pattanapisont2024}. Current methodologies lack standardised metrics to evaluate how 2D projections from different views distort or preserve critical gait dynamics. This gap motivates our development of a multi-metric framework (Dynamic Time Warping, Maximum Cross-Correlation, Kullback-Leibler Divergence, Information Entropy) to empirically validate camera view selection, transforming intuitive assumptions about frontal and lateral tradeoffs into actionable, quantitative guidelines.

The goal of the present work is to evaluate how the choice of recording view, frontal or lateral, affects the interpretation and characterisation of key kinematic and symmetry-based gait parameters in 2D gait analysis relative to 3D motion capture data. Using a multi-metric assessment framework, this study focuses on three key objectives:

1) Visualisation of Data Properties: Perform a comparative analysis of 2D gait data from frontal and lateral camera views alongside 3D motion capture data to uncover differences in data characteristics.

2) Impact of 2D Recording Views on Gait Parameters: Investigate how specific gait parameters, such as step length, knee rotation, distance between wrist and midpoint of hipline, and trunk rotation, are influenced by the choice of frontal or lateral views.

3) Optimal Camera View Selection: Develop strategies to identify the most suitable recording view for enhancing the clinical interpretability and reliability of gait analysis.

\section{Methods}
\begin{figure}[H]
\centering
\includegraphics[width=0.9\textwidth]{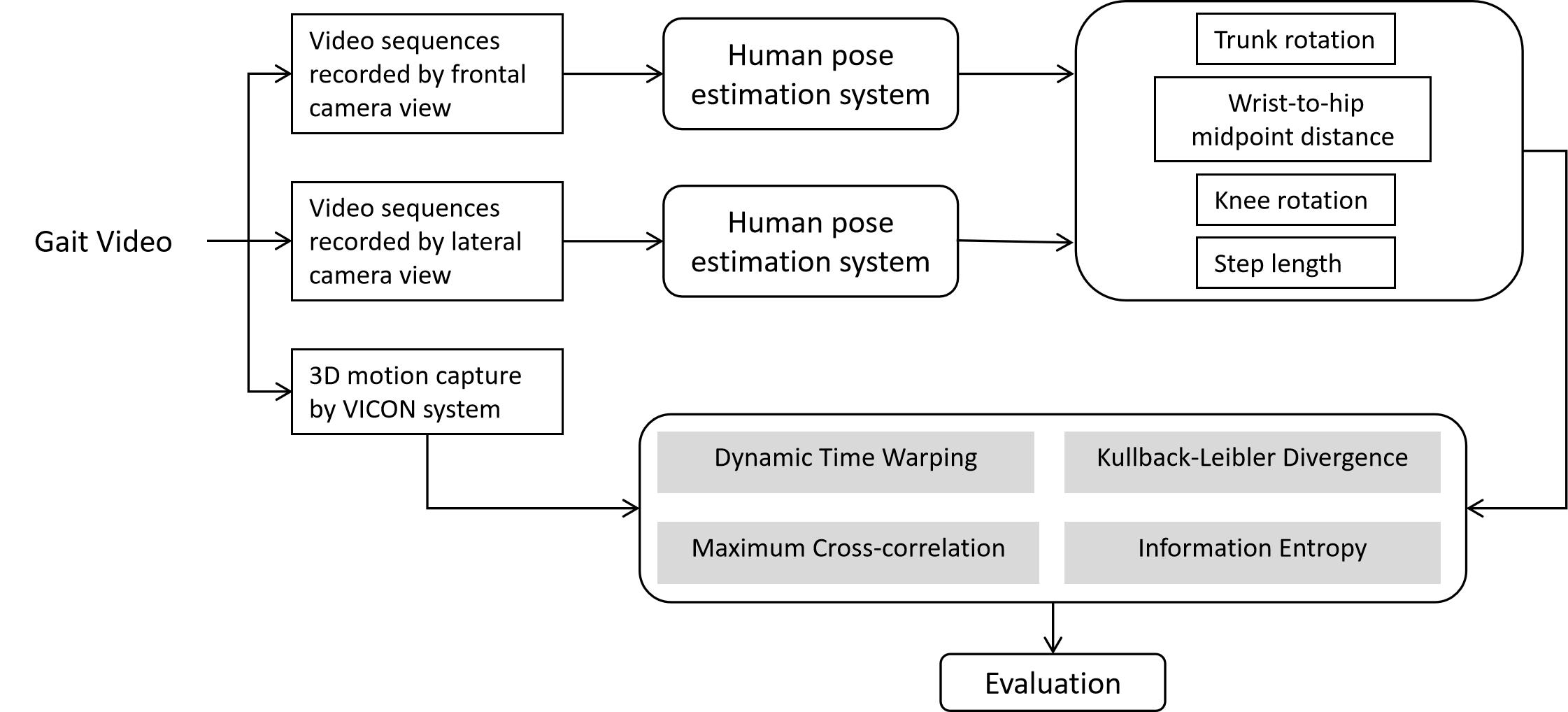} 
\caption{ Framework for multi-metric gait analysis using 2D projections. Pose estimation is performed using YOLOv8, producing frontal and lateral view projections. Gait parameters are extracted and analyzed using four metrics: Dynamic Time Warping (DTW), Maximum Cross-Correlation (MCC), Kullback-Leibler Divergence (KLD), and Information Entropy (IE). The results provide a comparative evaluation of camera views for specific gait factors.}
\label{fig:fig2}
\end{figure}

The proposed framework, as shown in Fig.~\ref{fig:fig2}, is for multi-metric gait analysis using 2D projections from frontal and lateral camera views. The workflow begins with the collection of gait videos recorded simultaneously from frontal and lateral views, alongside 3D motion capture data as the ground truth. Pose estimation is performed using the YOLOv8 framework to extract joint coordinates and motion trajectories from the 2D video recordings \cite{Dong2024}. 

To evaluate the quality and reliability of the 2D gait data, four metrics are summarized in Table~\ref{tab:metrics}, which contains Dynamic Time Warping (DTW) for assessing temporal alignment between 2D signals and 3D ground truth, Maximum Cross-Correlation (MCC) for measuring temporal synchronization and similarity between signals, Kullback-Leibler Divergence (KLD) for quantifying differences between probability distributions of 2D and 3D data, and Information Entropy (IE) for assessing the variability and complexity of motion patterns. These metrics enable a comparative analysis of frontal and lateral views, identifying the optimal camera views for specific gait parameters. This systematic framework provides a robust evaluation of 2D gait data, facilitating informed camera view selection for clinical and research applications.

\subsection*{A. Material Preparation}
Gait videos of 18 subjects were recorded at hospital using a VICON system with a camera rate of 100 Hz. The average number of frames per video was 169 ± 14. To ensure data quality, the signals were filtered using a zero-lag Butterworth filter with a cut-off frequency of 7 Hz and a sampling rate of 100 Hz. Frontal view recordings were obtained using a camera positioned 3 meters anterior to the walking path, elevated to 1.2 meters with a 10° downward tilt to optimise coronal plane visibility for parameters such as trunk rotation and hip-shoulder symmetry. Lateral view capture employed a camera placed 2.5 meters lateral at a height of 0.9 meters above the walking path. The YOLOv8 \cite{Dong2024} framework is utilized as a pose estimation system due to its convenient and accessible open-source API.

\subsection*{B. Applied Metrics}
To evaluate the quality and characteristics of gait data captured from different camera views (frontal and lateral) and their corresponding 2D projections, we consider the following framework shown in Fig.~\ref{fig:fig2}. The metrics are specifically selected for their ability to assess temporal alignment, similarity, and the informational content of data patterns:

\textbf{Dynamic Time Warping (DTW)} quantifies the temporal alignment between two time-series signals by measuring the optimal alignment of their trajectories \cite{Pattanapisont2024, Ozeloglu2024, Shah2025}. This metric is particularly useful for handling variations in walking speed, step duration, or other temporal irregularities, enabling robust comparisons between 2D projections and ground-truth 3D data. It ensures time-aligned evaluation despite differences in pace or step duration.

Let $\mathbf{x} = \{x_1, x_2, ..., x_n\}$ represent the 3D gait signal in step length, and $\mathbf{y} = \{y_1, y_2, ..., y_m\}$ represent the 2D gait signal in step length. The distance (typically Euclidean) between two points $x_i$ and $y_j$ is denoted as $d(x_i, y_j)$. The cumulative cost of aligning the first $i$ points of $\mathbf{x}$ with the first $j$ points of $\mathbf{y}$ is $D(i, j)$. The goal is to find the warping path $\mathcal{P}$, which defines the optimal alignment between the two sequences such that:
\begin{equation}
    D(i, j) = d(x_i, y_j) + \min \begin{cases}
        D(i-1, j) \\
        D(i, j-1) \\
        D(i-1, j-1)
    \end{cases}
\end{equation}

\textbf{Maximum Cross-Correlation (MCC)} measures the similarity between two signals as a function of the time lag applied to one signal relative to the other \cite{Lawin2023}. MCC identifies the highest correlation value, representing the greatest degree of similarity \cite{Patel2025, Hall2024}. Let $\mathbf{x}$ be the 3D gait signal and $\mathbf{y}$ be the 2D gait signal, and let $\tau$ be the time lag (shift) applied to one of the signals. The cross-correlation function $R_{xy}(\tau)$ is defined as:
\begin{equation}
    R_{xy}(\tau) = \sum_{t=1}^{N-\tau} x_t \cdot y_{t+\tau}
\end{equation}
where $N$ is the length of the signals. The maximum value of $R_{xy}(\tau)$ reflects the degree of similarity, while the lag $\tau$ at which it occurs indicates the temporal offset.

\textbf{Kullback-Leibler Divergence (KLD)} is a statistical measure that quantifies the difference between two probability distributions \cite{Hall2024, Chen2022}. In the context of gait analysis, KLD compares the distributions of extracted features (e.g., joint angles) from 2D projections with the corresponding 3D ground-truth data. Mathematically, KLD is defined as:
\begin{equation}
    D_{KL}(P \| Q) = \sum_{i} P(i) \log \frac{P(i)}{Q(i)}
\end{equation}
where $P$ is the probability distribution of the 3D gait signal, and $Q$ is the probability distribution of the 2D gait signal. A smaller $D_{KL}$ value indicates that $Q$ is closer to $P$.

\textbf{Information Entropy (IE)} is a statistical measure of uncertainty or randomness within a dataset \cite{Spineli2024, Lee2022, Zheng2024}. In gait analysis, IE quantifies the complexity or variability of motion patterns. Higher entropy values indicate greater diversity and richness in the data. For a dataset $X$ with possible events $x_i$, entropy is defined as:
\begin{equation}
    H(X) = -\sum_{i} p(x_i) \log p(x_i)
\end{equation}
where $p(x_i)$ is the probability of occurrence for event $x_i$ in the dataset.

\begin{table}[H]
\centering
\caption{Summary of metrics and functions}
\label{tab:metrics}
\begin{tabular}{ll}
\toprule
\textbf{Metrics} & \textbf{Functions} \\
\midrule
DTW & Temporal alignment \\
MCC & Similarity and correlation with the 3D ground truth \\
IE & Information retention and loss \\
KLD & Information retention and loss \\
\bottomrule
\end{tabular}
\caption*{\footnotesize * DTW: Dynamic Time Warping; MCC: Maximum Cross-Correlation; KLD: Kullback-Leibler Divergence; IE: Information Entropy.}
\end{table}

\subsection*{C. Feature Property}
\begin{figure}[H]
\centering
\includegraphics[width=0.9\textwidth]{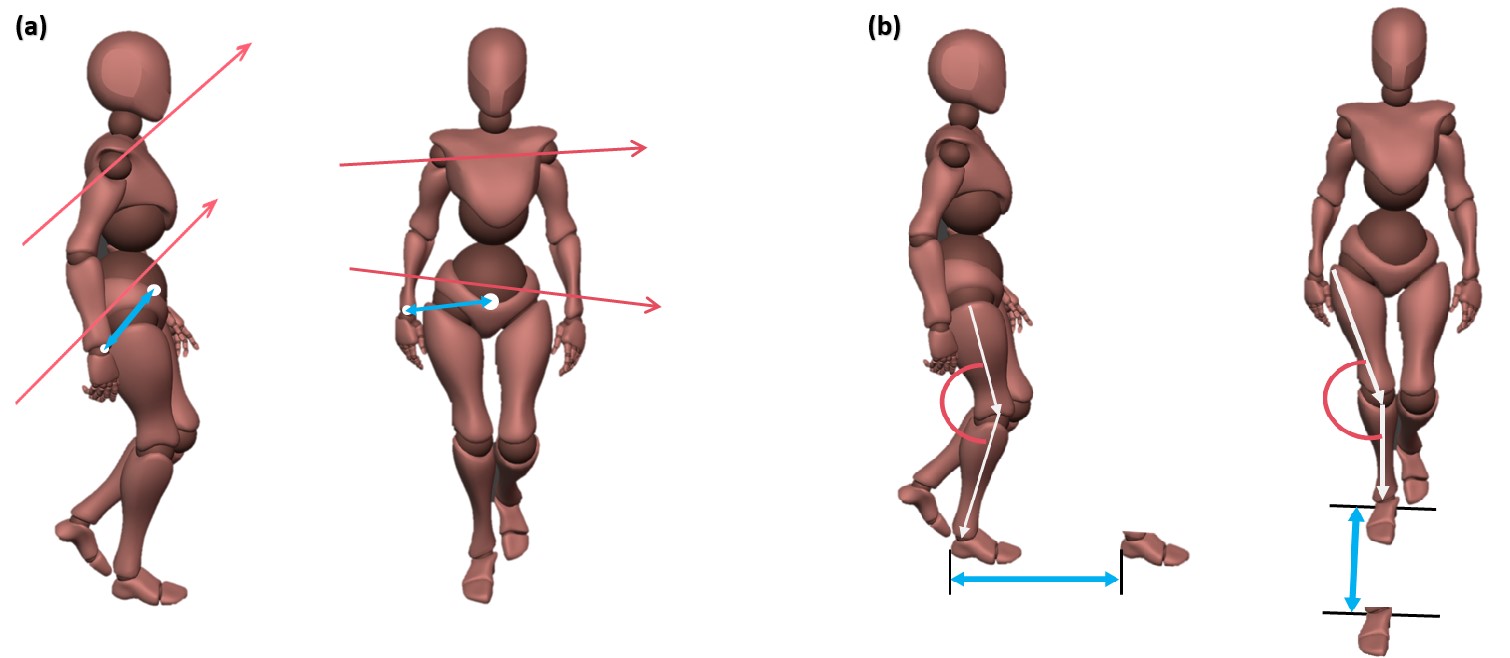} 
\caption{Visualization of gait measurement parameters. Red arrows indicate angle measurements, highlighting joint angles or angular trajectories between body segments. Blue arrows represent distance measurements, capturing spatial relationships such as step length or joint displacements. a) These features calculated using landmarks on a both sides. b) These features calculated using landmarks on a single side.}
\label{fig:fig3}
\end{figure}

The selection of factors for this study was guided by the need to effectively compare 2D signals from frontal and lateral camera views with corresponding 3D motion capture data. To achieve this, we focused on four kinematic features that are representative of gait dynamics, categorized into two types based on how the features are calculated from anatomical landmarks, as shown in Fig.~\ref{fig:fig3}.

Features of \textbf{single side} include \textbf{step length}, which quantifies the spatial progression of gait and is derived from the displacement of foot landmarks during walking; and \textbf{knee rotation}, which captures the angular motion of the knee joint, providing insights into the angular dynamics within a single limb during gait.

Features of \textbf{bilateral sides} include \textbf{trunk rotation}, which measures the rotational movement of the upper body during gait, offering valuable information about overall body coordination; and \textbf{distance between wrist and midpoint of hip}, which is calculated as the distance between the wrist and the midpoint between the left and right hip joints, highlighting the arm swing dynamics relative to the pelvis, providing additional context about upper-body motion during walking.

By selecting these features, we aimed to capture both localized joint-level movements (e.g., step length and knee rotation) and more global, coordinated body movements (e.g., trunk rotation and wrist-to-midpoint of hipline). This dual approach provides a comprehensive basis for comparing gait dynamics across different data capture modalities, ensuring that both individual limb movements and overall body coordination are effectively represented. These features were specifically chosen for their relevance to gait analysis and their ability to highlight the differences and similarities between 2D and 3D data.

\subsection*{D. Statistical Analysis}
To determine whether observed differences in metric scores (DTW, MCC, KLD, IE) between frontal and lateral views were statistically significant, we performed non-parametric Wilcoxon signed-rank tests. This test was chosen due to the small sample size ($n = 18$) and non-normal distribution of metric scores (confirmed via Shapiro-Wilk tests). Effect sizes were calculated using Cliff’s delta ($\delta$), a robust measure for non-parametric data, with thresholds: $\delta < 0.33$ (small), $0.33 \leq \delta < 0.474$ (medium), $\delta \geq 0.474$ (large). Analyses were conducted in Python using SciPy (v1.13.0) and NumPy (v1.26.4).

\section{Results}
\subsection{Dimensionality Reduction for Gait Data}

To compare gait data recorded from different perspectives, such as frontal camera view, lateral camera view, and 3D motion capture systems, we first reduced the dimensionality of the datasets using Principal Component Analysis (PCA). A 95\% variance threshold was applied to retain most of the variability in the data while minimising redundancy. The results of the dimensionality reduction are summarised in Table~\ref{tab:dimensionality_reduction}.

\begin{table}[H]
\centering
\caption{Dimensionality Reduction Results of Each Data Group}
\label{tab:dimensionality_reduction}
\begin{tabular}{lccc}
\toprule
\textbf{Metrics} & \textbf{Initial Dimension} & \textbf{95\% PCA Dimension} & \textbf{Explained Variance} \\
\midrule
Frontal & 34 & 6 & 95.9\% \\
Lateral & 34 & 13 & 95.5\% \\
3D motion capture & 40 & 10 & 95.6\% \\
\bottomrule
\end{tabular}
\end{table}

The backward camera view gait data, initially consisting of 34 features, was reduced to 6 principal components, capturing an explained variance ratio of 95.9\%. This indicates that the reduced dataset retains nearly all the variance of the original data while significantly reducing the feature space.
For the lateral camera view, the initial 34 dimensional data was reduced to 13 principal components, with an explained variance ratio of 95.5\%. The higher number of principal components required to achieve the 95\% variance threshold compared to the backward view suggests greater complexity or variability in the lateral view data.
The 3D motion capture system, containing 40 original features, was reduced to 10 principal components, achieving an explained variance ratio of 95.6\%. This reduction highlights the efficiency of PCA in compressing high-dimensional 3D data while preserving essential gait dynamics.

\subsection{Visualisation of Data Patterns}
\begin{figure}[H]
\centering
\includegraphics[width=0.9\textwidth]{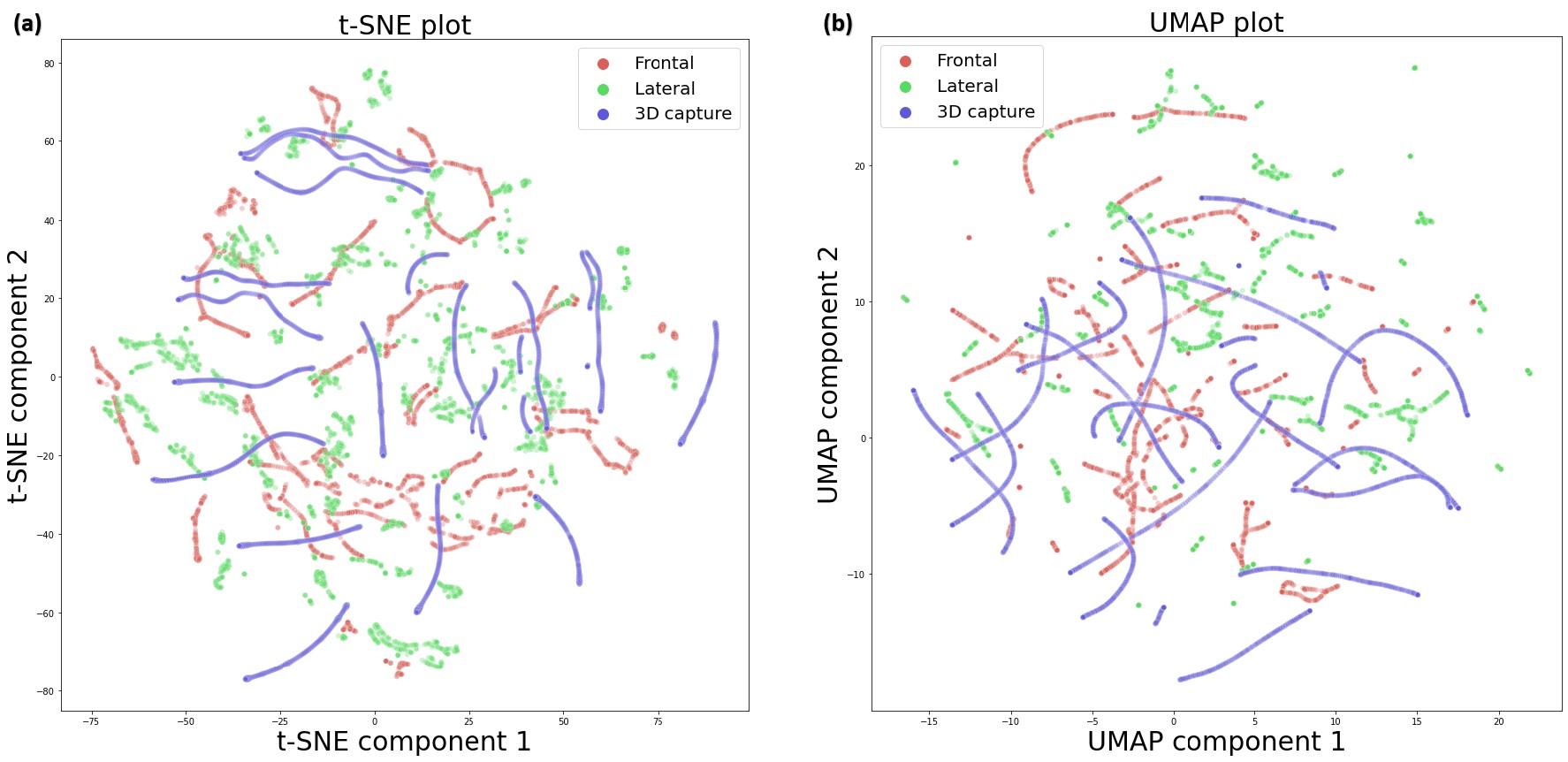} 
\caption{ (a) t-SNE visualization of gait data from frontal and lateral camera views and 3D motion capture systems. Red points represent the frontal view, green points the lateral view, and blue points the 3D motion capture data. (b) UMAP visualization of gait data captured from frontal and lateral camera views, as well as 3D motion capture systems. Red points represent data from the frontal view, green points represent the lateral view, and blue points correspond to 3D motion capture data. The clustering and dispersion patterns illustrate similarities between the two 2D camera views and the broader variability captured by the 3D motion data, highlighting its ability to represent more complex gait dynamics.}
\label{fig:fig4}
\end{figure}

To study the differences in gait data recorded from the frontal and lateral camera views, as well as the 3D motion capture system, we applied t-distributed Stochastic Neighbor Embedding (t-SNE) and Uniform Manifold Approximation and Projection (UMAP) to visualize the 95\% PCA data patterns in a lower-dimensional space, as shown in Fig.~\ref{fig:fig4}.

Both t-SNE and UMAP visualizations reveal distinct clustering patterns among the data derived from frontal and lateral cameras (represented by red and green points) and 3D motion capture systems (purple points). Specifically, the data from the frontal and lateral cameras tend to cluster closely, suggesting a notable similarity in the gait information captured from these two views. In contrast, the 3D motion capture data exhibits greater dispersion, reflecting its ability to capture a broader range of variability in gait patterns. This increased variability is likely due to the depth information and complex movements detected by 3D motion capture systems, which are not as evident in 2D video data.

The partial overlap between the frontal and lateral camera data points indicates a degree of redundancy in the information captured from these two views. This suggests that, for certain analyses, data from a single camera view may suffice without a substantial loss of information. However, the differences between the 2D camera data and the 3D motion capture data highlight the complementary nature of these methods: while 2D video primarily captures planar motion, 3D motion capture provides a more comprehensive representation of gait, including depth and intricate movement patterns.

\subsection{Multi-Metric Evaluation of Key Gait Parameters}
The analysis of key gait parameters was conducted using multi-metric approach, incorporating Dynamic Time Warping (DTW), Maximum Cross-Correlation (MCC), Information Entropy (IE), and Kullback-Leibler Divergence (KLD). These metrics were used to evaluate the temporal alignment, correlation, variability, and similarity of gait joint motion between the frontal and lateral camera views, relative to the 3D ground truth. 

\subsubsection{Trunk Rotation}
\begin{figure}[H]
\centering
\includegraphics[width=0.7\textwidth]{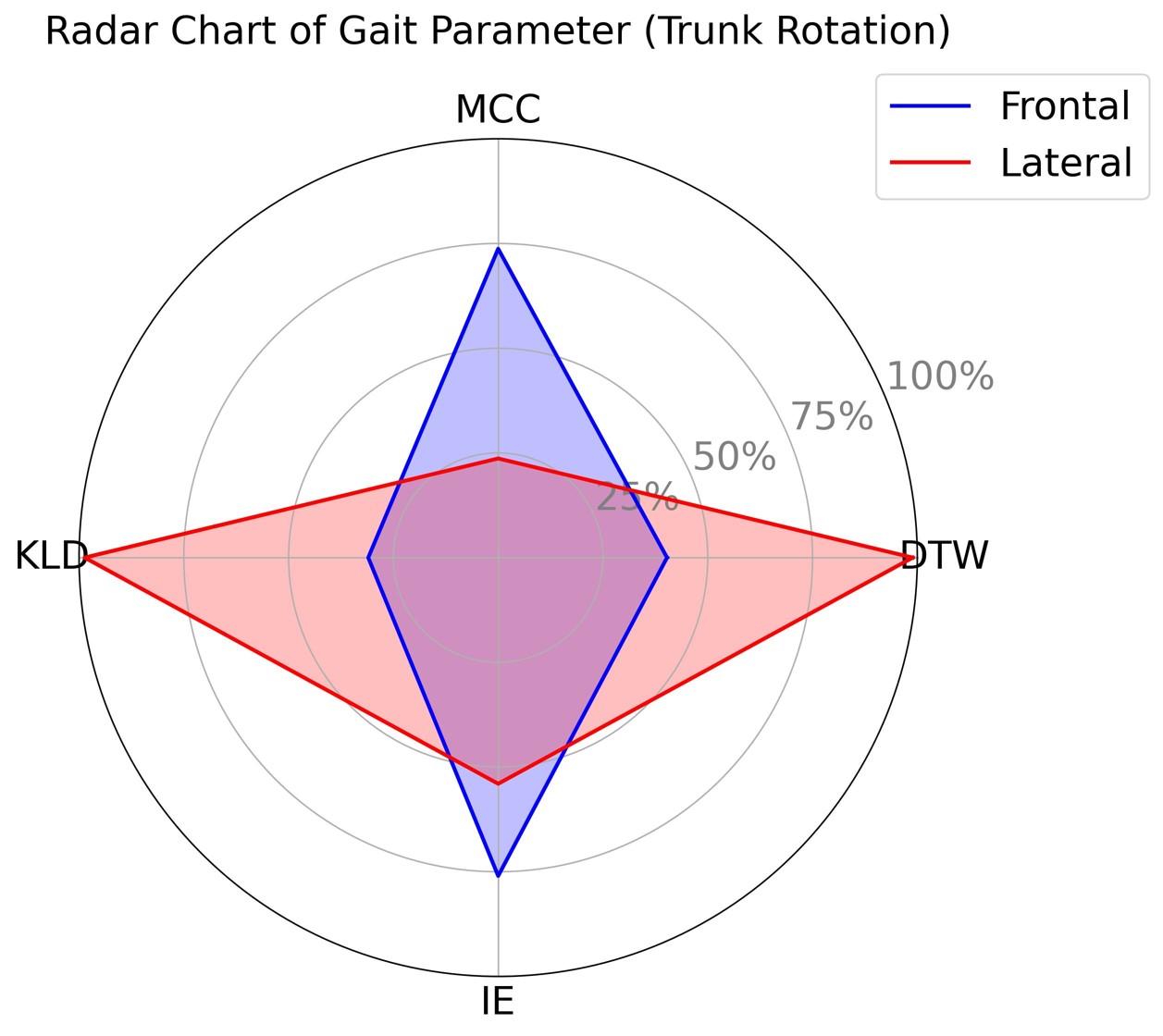} 
\caption{ Radar chart comparing the performance of frontal and lateral views in 3D motion capture projection of trunk rotation analysis across four metrics}
\label{fig:fig5}
\end{figure}

As illustrated in Fig.~\ref{fig:fig5} and detailed in Table~\ref{tab:trunk_rotation_radar}, the frontal view demonstrated superior alignment and fidelity to 3D motion capture ground truth across all metrics, though statistical significance varied.

For temporal alignment (DTW), the frontal view showed a mean value of $99.03 \pm 21.81$, compared to $104.89 \pm 17.45$ for the lateral view. The frontal view exhibited lower DTW values (indicating better alignment), though this difference was not statistically significant ($p = 0.702$; Cliff's $\delta = 0.14$, small effect size). Similarly, MCC values were marginally higher in the frontal view ($71.11 \pm 20.50$ vs. $70.35 \pm 21.85$ for lateral), but this difference was also non-significant ($p = 0.865$; Cliff's $\delta = 0.08$, small effect size). Information entropy (IE) values were slightly higher in the frontal view ($7.15 \pm 0.23$ vs. $7.11 \pm 0.18$), suggesting a more distributed representation of movement dynamics, though this difference remained non-significant ($p = 0.119$; Cliff's $\delta = 0.13$, small effect size).

The most pronounced difference emerged in KLD, which quantifies distributional similarity to 3D rotation. The frontal view achieved a significantly lower KLD ($0.09 \pm 0.06$) compared to the lateral view ($0.30 \pm 0.19$; $p < 0.001$, Cliff's $\delta = 0.76$, large effect size). This indicates that the frontal projection retained rotational measurements more accurately, aligning closely with the 3D ground truth. The radar chart (Fig.~\ref{fig:fig5}) visually reinforces these findings, showing the frontal view's consistent advantage across metrics, particularly in KLD.

\begin{table}[H]
\centering
\caption{Comparison of 3D motion capture metrics for trunk rotation across different projection views}
\label{tab:trunk_rotation_radar}
\begin{tabular}{lcccc}
\toprule
\textbf{Metrics} & \textbf{Frontal} & \textbf{Lateral} & \textbf{P-value} & \textbf{Cliff's $\delta$} \\
\midrule
DTW & $99.03 \pm 21.81$ & $104.89 \pm 17.45$ & $0.702$ & $0.14$ (Small) \\
MCC & $71.11 \pm 20.50$ & $70.35 \pm 21.85$ & $0.865$ & $0.08$ (Small) \\
IE & $7.15 \pm 0.23$ & $7.11 \pm 0.18$ & $0.119$ & $0.13$ (Small) \\
KLD & $0.09 \pm 0.06$ & $0.30 \pm 0.19$ & $<0.001$ & $0.76$ (Large) \\
\bottomrule
\end{tabular}

\footnotesize{* $p<0.05$; thresholds: small ($\delta<0.33$), medium ($0.33 \leq \delta<0.474$), large ($\delta \geq 0.474$); DTW: Dynamic Time Warping; MCC: Maximum Cross-Correlation; KLD: Kullback-Leibler Divergence; IE: Information Entropy}
\end{table}

\subsubsection{Distance between Wrist and Midpoint of Hipline}
\begin{figure}[H]
\centering
\includegraphics[width=0.9\textwidth]{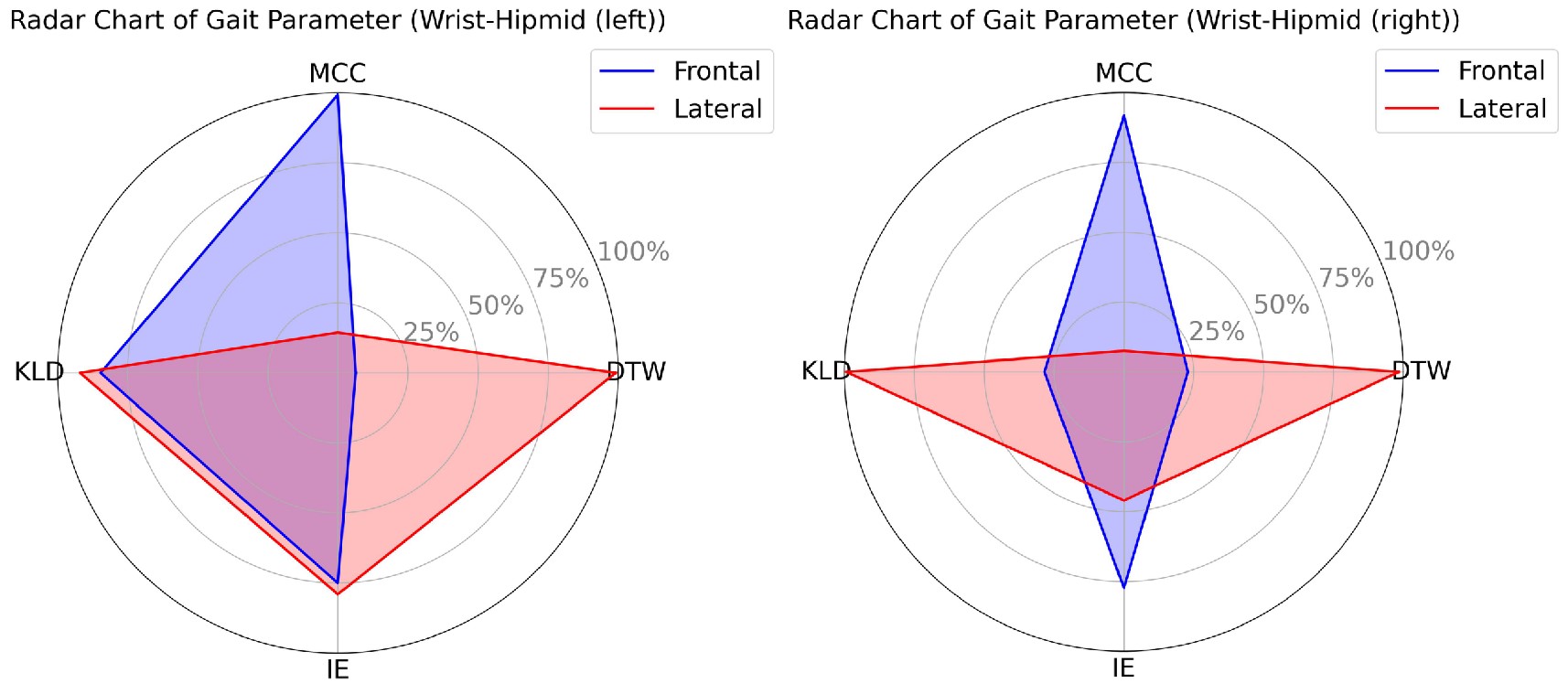} 
\caption{ Radar chart comparing frontal and lateral view performance in 3D motion capture analysis, measuring the distance between the wrists and the hip joint midpoint across four key metrics}
\label{fig:fig6}
\end{figure}

The performance of frontal and lateral 2D projections in capturing wrist-to-hipmid dynamics was evaluated as shown in Fig.~\ref{fig:fig6} and Table~\ref{tab:wrist_hipmid}. For the left wrist-to-hipmid analysis, the frontal view demonstrated significantly lower DTW ($62.90 \pm 23.55$ vs. $105.59 \pm 38.00$, $p = 0.003$, Cliff's $\delta = 0.63$, large effect size) and higher MCC ($105.77 \pm 29.72$ vs. $75.20 \pm 20.38$, $p = 0.003$, Cliff's $\delta = 0.58$, large effect size) compared to the lateral view, indicating superior temporal alignment and correlation with 3D ground truth. Information entropy (IE) values were marginally higher in the lateral view ($7.24 \pm 0.15$ vs. $7.23 \pm 0.21$), though this difference was non-significant ($p = 0.799$, Cliff's $\delta = 0.01$, small effect size). KLD values showed no significant difference between views ($p = 0.495$, Cliff's $\delta = 0.24$, small effect size), though the frontal view retained a slight advantage ($0.13 \pm 0.16$ vs. $0.14 \pm 0.10$).

For the right wrist-to-hipmid analysis, the frontal view achieved significantly higher MCC ($97.12 \pm 35.73$ vs. $67.64 \pm 20.23$, $p = 0.009$, Cliff's $\delta = 0.46$, medium effect size) and lower KLD ($0.13 \pm 0.13$ vs. $0.45 \pm 0.64$, $p = 0.003$, Cliff's $\delta = 0.59$, large effect size) than the lateral view, reflecting stronger correlation and more accurate retention of rotational measurements. DTW values were lower in the frontal view ($75.50 \pm 40.44$ vs. $93.62 \pm 25.91$), though non-significant ($p = 0.119$, Cliff's $\delta = 0.35$, medium effect size). Notably, the frontal view exhibited significantly lower IE values ($7.31 \pm 0.15$ vs. $7.18 \pm 0.21$, $p = 0.004$, Cliff's $\delta = 0.34$, medium effect size), suggesting reduced motion variability compared to the lateral view.

Radar charts (Fig.~\ref{fig:fig6}) visually confirmed the frontal view's consistent superiority across metrics, particularly in DTW, MCC, and KLD for both sides. While the lateral view captured greater motion variability (higher IE), its inferior alignment and correlation with 3D ground truth underscored the frontal view's reliability for upper-body motion analysis.

\begin{table}[H]
\centering
\caption{Comparison of 3D motion capture metrics for distance between the wrists and the hip joint midpoint in different projection views}
\label{tab:wrist_hipmid}
\begin{tabular}{lcccc}
\toprule
\textbf{Metrics} & \textbf{Frontal} & \textbf{Lateral} & \textbf{P-value} & \textbf{Cliff's $\delta$} \\
\midrule
DTW (L) & $62.90 \pm 23.55$ & $105.59 \pm 38.00$ & $0.003$ & $0.63$ (Large) \\
MCC (L) & $105.77 \pm 29.72$ & $75.20 \pm 20.38$ & $0.003$ & $0.58$ (Large) \\
IE (L) & $7.23 \pm 0.21$ & $7.24 \pm 0.15$ & $0.799$ & $0.01$ (Small) \\
KLD (L) & $0.13 \pm 0.16$ & $0.14 \pm 0.10$ & $0.495$ & $0.24$ (Small) \\
DTW (R) & $75.50 \pm 40.44$ & $93.62 \pm 25.91$ & $0.119$ & $0.35$ (Medium) \\
MCC (R) & $97.12 \pm 35.73$ & $67.64 \pm 20.23$ & $0.009$ & $0.46$ (Medium) \\
IE (R) & $7.31 \pm 0.15$ & $7.18 \pm 0.21$ & $0.004$ & $0.34$ (Medium) \\
KLD (R) & $0.13 \pm 0.13$ & $0.45 \pm 0.64$ & $0.003$ & $0.59$ (Large) \\
\bottomrule
\end{tabular}

\footnotesize{* $p<0.05$; thresholds: small ($\delta<0.33$), medium ($0.33 \leq \delta<0.474$), large ($\delta \geq 0.474$); DTW: Dynamic Time Warping; MCC: Maximum Cross-Correlation; KLD: Kullback-Leibler Divergence; IE: Information Entropy; L represents left side, R represents right side.}
\end{table}

\subsubsection{Step Length}
\begin{figure}[H]
\centering
\includegraphics[width=0.9\textwidth]{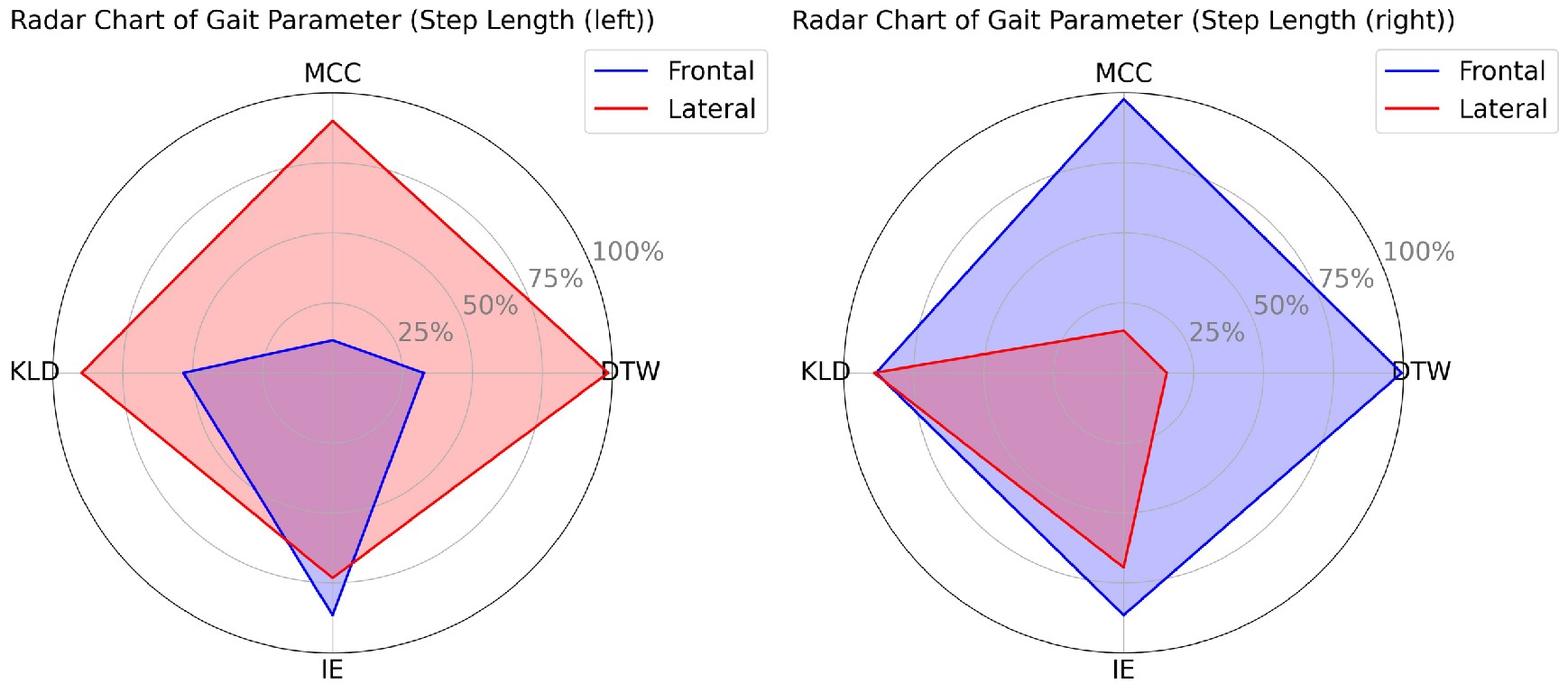} 
\caption{Radar chart comparing frontal and lateral view performance in 3D motion capture analysis, measuring step length across four key metrics}
\label{fig:fig7}
\end{figure}

The performance of frontal and lateral 2D projections in capturing step length dynamics was evaluated as illustrated in Fig.~\ref{fig:fig7} and Table~\ref{tab:step_length}. The lateral view demonstrated advantages in temporal alignment for the right side, while the frontal view excelled in preserving spatial features critical for gait analysis.

For the left step length, the frontal view achieved a mean DTW of $98.58 \pm 11.21$ and KLD of $0.16 \pm 0.07$, compared to the lateral view's DTW of $105.83 \pm 22.77$ and KLD of $0.27 \pm 0.15$. While the lateral view showed marginally higher MCC scores ($76.61 \pm 24.48$ vs. $65.63 \pm 14.42$; $p = 0.054$, Cliff's $\delta = 0.34$, medium effect size), the frontal view achieved significantly lower KLD values ($p = 0.004$, Cliff's $\delta = 0.45$, medium effect size), indicating better spatial alignment with 3D ground truth. Information entropy (IE) values were comparable (frontal: $7.35 \pm 0.13$; lateral: $7.29 \pm 0.16$; $p = 0.064$, Cliff's $\delta = 0.15$, small effect size).

For the right step length, the lateral view outperformed the frontal view in temporal alignment, achieving a lower DTW ($53.08 \pm 24.50$ vs. $69.87 \pm 25.36$; $p = 0.005$, Cliff's $\delta = 0.41$, medium effect size) and slightly higher IE values ($7.28 \pm 0.16$ vs. $7.35 \pm 0.12$; $p = 0.019$, Cliff's $\delta = 0.22$, small effect size). MCC scores were similar (lateral: $123.12 \pm 34.93$; frontal: $123.78 \pm 25.54$; $p = 0.609$, Cliff's $\delta = 0.07$, small effect size), and KLD values showed no significant difference (lateral: $0.17 \pm 0.13$; frontal: $0.18 \pm 0.15$; $p = 0.966$, Cliff's $\delta = 0.25$, small effect size).

Radar charts (Fig.~\ref{fig:fig7}) visually reinforced these trends: the lateral view excelled in temporal metrics (e.g., right DTW), while the frontal view dominated in spatial metrics (e.g., left KLD). This suggests that the lateral view's reduced occlusion and measurement errors enhance temporal accuracy on walking side, whereas the frontal view's broader field of view preserves spatial features critical for gait analysis.

\begin{table}[H]
\centering
\caption{Comparison of 3D motion capture metrics for step length in different projection views}
\label{tab:step_length}
\begin{tabular}{lcccc}
\toprule
\textbf{Metrics} & \textbf{Frontal} & \textbf{Lateral} & \textbf{P-value} & \textbf{Cliff's $\delta$} \\
\midrule
DTW (L) & $98.58 \pm 11.21$ & $105.83 \pm 22.77$ & $0.246$ & $0.20$ (Small) \\
MCC (L) & $65.63 \pm 14.42$ & $76.61 \pm 24.48$ & $0.054$ & $0.34$ (Medium) \\
IE (L) & $7.35 \pm 0.13$ & $7.29 \pm 0.16$ & $0.064$ & $0.15$ (Small) \\
KLD (L) & $0.16 \pm 0.07$ & $0.27 \pm 0.15$ & $0.004$ & $0.45$ (Medium) \\
DTW (R) & $69.87 \pm 25.36$ & $53.08 \pm 24.50$ & $0.005$ & $0.41$ (Medium) \\
MCC (R) & $123.78 \pm 25.54$ & $123.12 \pm 34.93$ & $0.609$ & $0.07$ (Small) \\
IE (R) & $7.35 \pm 0.12$ & $7.28 \pm 0.16$ & $0.019$ & $0.22$ (Small) \\
KLD (R) & $0.12 \pm 0.04$ & $0.12 \pm 0.10$ & $0.966$ & $0.25$ (Small) \\
\bottomrule
\end{tabular}

\footnotesize{* $p<0.05$; thresholds: small ($\delta<0.33$), medium ($0.33 \leq \delta<0.474$), large ($\delta \geq 0.474$); DTW: Dynamic Time Warping; MCC: Maximum Cross-Correlation; KLD: Kullback-Leibler Divergence; IE: Information Entropy; L represents left side, R represents right side.}
\end{table}

\subsubsection{Knee Rotation}
\begin{figure}[H]
\centering
\includegraphics[width=0.9\textwidth]{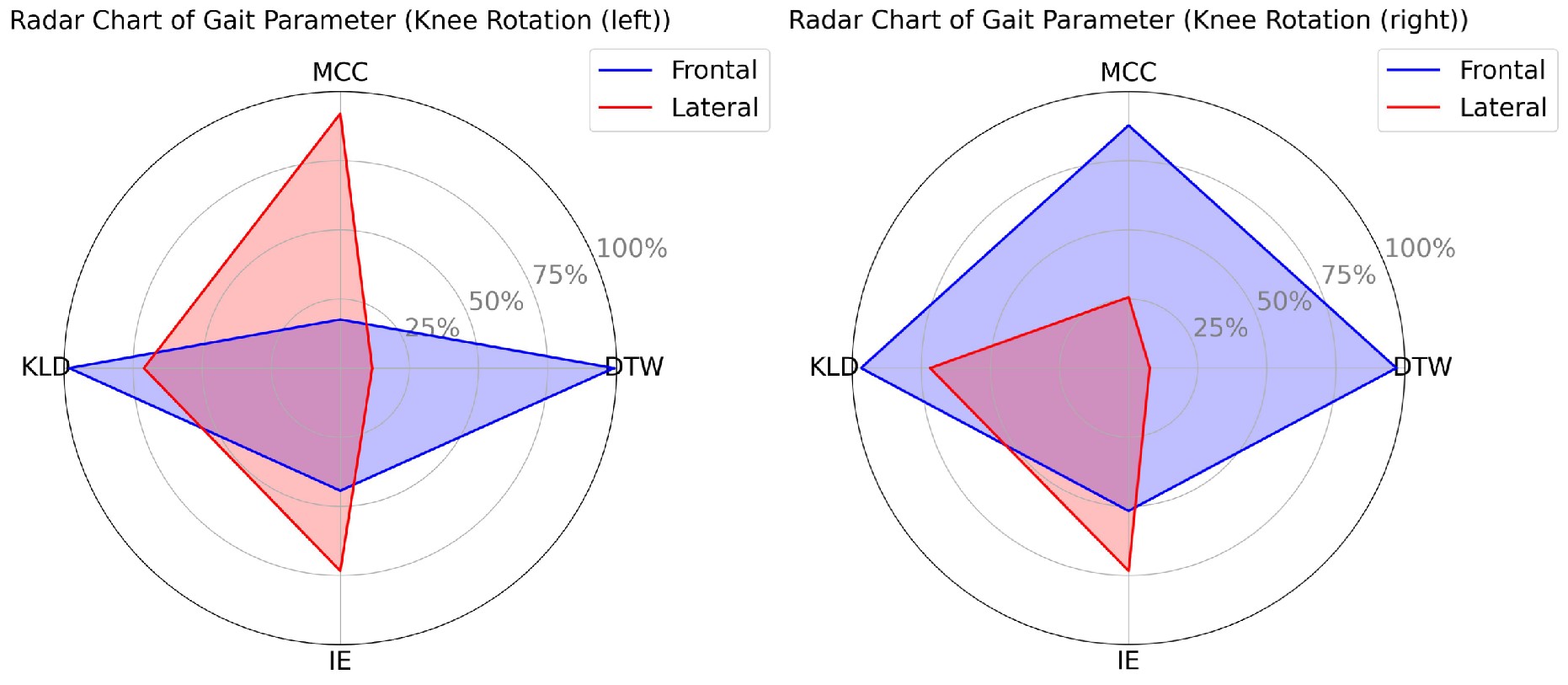} 
\caption{Radar chart comparing frontal and lateral view performance in 3D motion capture analysis, measuring step length across four key metrics}
\label{fig:fig8}
\end{figure}

The performance of frontal and lateral 2D projections in capturing knee rotation dynamics was assessed as shown in Fig.~\ref{fig:fig8} and Table~\ref{tab:knee_rotation}. The lateral view demonstrated superior alignment and similarity to 3D ground truth, particularly in temporal metrics (DTW), while the frontal view retained spatial features critical for rotational measurements.

For left knee rotation, the lateral view achieved significantly lower DTW values ($106.46 \pm 38.57$ vs. $155.41 \pm 41.77$; $p = 0.004$, Cliff's $\delta = 0.62$, large effect size), indicating better temporal alignment. MCC scores were comparable between views (lateral: $79.15 \pm 38.65$; frontal: $63.66 \pm 15.36$; $p = 0.523$, Cliff's $\delta = 0.20$, small effect size). However, the frontal view showed significantly higher IE values ($7.33 \pm 0.13$ vs. $7.42 \pm 0.12$; $p < 0.001$, Cliff's $\delta = 0.36$, medium effect size), suggesting greater variability in motion patterns. KLD values were lower in the lateral view ($1.86 \pm 1.09$ vs. $1.35 \pm 0.99$; $p = 0.167$, Cliff's $\delta = 0.24$, small effect size), though this difference was non-significant.

For right knee rotation, the lateral view again exhibited lower DTW values ($93.41 \pm 28.09$ vs. $76.46 \pm 41.04$; $p = 0.099$, Cliff's $\delta = 0.19$, small effect size), though this difference approached significance. MCC scores were similar (lateral: $93.02 \pm 40.40$; frontal: $100.36 \pm 31.02$; $p = 0.671$, Cliff's $\delta = 0.17$, small effect size). The lateral view also showed slightly higher IE values ($7.35 \pm 0.13$ vs. $7.42 \pm 0.12$; $p < 0.001$, Cliff's $\delta = 0.33$, small effect size), while KLD values remained comparable ($p = 0.369$, Cliff's $\delta = 0.21$, small effect size).

Radar charts (Fig.~\ref{fig:fig8}) visually highlighted the lateral view's dominance in DTW and MCC for both sides, whereas the frontal view retained higher IE values, reflecting its utility in capturing rotation. These findings suggest that the lateral view's proximity to the camera reduces measurement errors, enhancing temporal alignment, while the frontal view's broader field of view preserves spatial details critical for rotational analysis.

\begin{table}[H]
\centering
\caption{Comparison of 3D motion capture metrics for knee rotation in different projection views}
\label{tab:knee_rotation}
\begin{tabular}{lcccc}
\toprule
\textbf{Metrics} & \textbf{Frontal} & \textbf{Lateral} & \textbf{P-value} & \textbf{Cliff's $\delta$} \\
\midrule
DTW (L) & $155.41 \pm 41.77$ & $106.46 \pm 38.57$ & $0.004$ & $0.62$ (Large) \\
MCC (L) & $63.66 \pm 15.36$ & $79.15 \pm 38.65$ & $0.523$ & $0.20$ (Small) \\
IE (L) & $7.33 \pm 0.13$ & $7.42 \pm 0.12$ & $<0.001$ & $0.36$ (Medium) \\
KLD (L) & $1.86 \pm 1.09$ & $1.35 \pm 0.99$ & $0.167$ & $0.24$ (Small) \\
DTW (R) & $93.41 \pm 28.09$ & $76.46 \pm 41.04$ & $0.099$ & $0.19$ (Small) \\
MCC (R) & $100.36 \pm 31.02$ & $93.02 \pm 40.40$ & $0.671$ & $0.17$ (Small) \\
IE (R) & $7.35 \pm 0.13$ & $7.42 \pm 0.12$ & $<0.001$ & $0.33$ (Small) \\
KLD (R) & $1.07 \pm 1.02$ & $0.79 \pm 0.77$ & $0.369$ & $0.21$ (Small) \\
\bottomrule
\end{tabular}

\footnotesize{* $p<0.05$; thresholds: small ($\delta<0.33$), medium ($0.33 \leq \delta<0.474$), large ($\delta \geq 0.474$); DTW: Dynamic Time Warping; MCC: Maximum Cross-Correlation; KLD: Kullback-Leibler Divergence; IE: Information Entropy; L represents left side, R represents right side.}
\end{table}

\section{Discussion}
This study advances markerless gait analysis by introducing a multi-metric evaluation framework (DTW, MCC, KLD, IE) to address the clinical dilemma of selecting optimal camera views for specific gait parameters. Although prior studies have suggested that lateral views better capture sagittal kinematics \cite{Pattanapisont2024, Slembrouck2020}, this work extends these findings by providing three novel capabilities. First, it quantifies view-specific information loss using metrics like KLD and IE, moving beyond traditional correlation measures. Second, it identifies the optimal view for each parameter through cross-metric consensus, enabling parameter-specific guidance. Third, it establishes benchmark values for future system validation, bridging qualitative clinical intuition with evidence-based decision-making. These advancements are particularly relevant for institutions transitioning from costly 3D systems to cost-effective 2D setups.

The study's findings demonstrate clear distinctions between frontal and lateral views based on the gait parameter being analyzed. Lateral views proved superior for sagittal plane motion parameters such as step length and knee rotation. This advantage is attributed to reduced occlusion and clearer joint visibility in the walking direction, which enhances temporal alignment (DTW) and signal similarity (MCC). These results align with prior research emphasizing the importance of sagittal plane visibility for accurate lower-limb kinematics \cite{Pattanapisont2024, Slembrouck2020}. Conversely, frontal views excelled in analyzing rotational and symmetry-based parameters, such as trunk rotation and wrist-to-hipmid distances. The frontal view provided stronger correlations and retained critical information for assessing balance disorders or asymmetrical gait patterns, making it particularly valuable for conditions involving coronal plane dynamics \cite{Okuma2014, daSilveira2022}. Statistical validation confirmed that differences between views were both statistically significant and clinically meaningful. For instance, lateral views showed significantly lower DTW values for right step length ($p = 0.005$, $\delta = 0.41$, medium effect size) and marginally higher MCC scores for left step length ($p = 0.054$, $\delta = 0.34$, medium effect size). In contrast, frontal views achieved significantly lower KLD values for trunk rotation ($p < 0.001$), reflecting better retention of rotation. These effect sizes highlight the critical role of camera view selection in ensuring gait parameter fidelity.

The practical implications of these findings are substantial. Selecting the appropriate camera view depends on the specific gait parameter being analysed. Lateral views are recommended for lower-body kinematics, such as step length and knee rotation, due to their alignment with sagittal plane motion. Frontal views, on the other hand, are more suitable for upper-body symmetry and trunk rotation assessments, particularly in patients with balance disorders or asymmetrical gait patterns \cite{Stenum2024, Pattanapisont2024, Lawin2023}. Multi-camera setups could further enhance accuracy by combining the strengths of both views, mitigating occlusion and noise in real-world environments. Such an approach would enable clinicians and researchers to achieve a more comprehensive understanding of gait dynamics, particularly in scenarios where variability and environmental factors pose challenges.

Methodologically, this study addresses key limitations of prior single-metric approaches. Earlier works often relied solely on correlation coefficients or visual pattern matching \cite{Stenum2021, Pattanapisont2024}, which fail to capture the trade-offs between views. By integrating multiple metrics---DTW for temporal alignment, MCC for signal similarity, KLD for distributional fidelity, and IE for information content---this framework reveals previously unreported nuances. For example, lateral views exhibited superior DTW scores for step length but higher KLD values for wrist-to-hipmid distances, a contradiction resolvable only through multi-metric analysis. This granular approach provides parameter-specific guidance that single-view or single-metric studies cannot achieve, offering a more robust foundation for clinical decision-making.

Despite these contributions, the study has limitations. The statistical power is constrained by the cohort size ($n = 18$), necessitating larger, more diverse datasets to generalize effect sizes across populations. Additionally, reliance on side-specific measurements introduces variability, particularly in lateral views where occlusion can skew results for certain parameters. The use of KLD and IE highlights differences in information retention and complexity but may not fully capture the dynamic interdependencies between 2D views and 3D motion, limiting the ability to model complex gait dynamics accurately. Future work should explore advanced methodologies for fusing multi-view data and developing more robust metrics to address these gaps.

Promising directions for future research include the development of advanced metrics to better model gait dynamics across perspectives. Incorporating temporal-spatial coherence metrics or machine learning-based approaches could enable more nuanced evaluations of gait patterns. Furthermore, validating these findings in specific clinical populations, such as individuals with stroke, cerebral palsy, or degenerative joint disorders, could provide valuable insights into the utility of markerless systems for diagnosing and monitoring gait abnormalities. By addressing these challenges and building on the current framework, future studies can further refine markerless gait analysis, enhancing its reliability and applicability in clinical and real-world settings.

\section{Conclusion}
This study presents a systematic, multi-metric evaluation framework to guide camera view selection in markerless 2D gait analysis. By comparing frontal and lateral views against 3D motion capture ground truth using Dynamic Time Warping (DTW), Maximum Cross-Correlation (MCC), Kullback-Leibler Divergence (KLD), and Information Entropy (IE), we provide evidence-based recommendations for optimizing view selection based on the target gait parameter.

Our results reveal a clear performance dichotomy: lateral views are superior for sagittal plane kinematics, such as step length and knee rotation, demonstrating significantly better temporal alignment (lower DTW, $p < 0.01$, medium-to-large effect sizes). Conversely, frontal views excel in capturing symmetry and rotational dynamics, particularly trunk rotation and arm swing (wrist-to-hipmid distance), where they show significantly lower KLD values ($p < 0.001$), indicating a more accurate representation of spatial distributions.

These findings establish that there is no universally optimal camera view. Instead, the choice must be parameter-specific. For clinical assessments focused on lower-limb mechanics and temporal gait patterns, a lateral view is recommended. For evaluations of postural symmetry, balance, and upper-body coordination, a frontal view is preferred. This work transforms camera selection from an intuitive decision into a data-driven process, significantly enhancing the accuracy and reliability of 2D markerless systems. The proposed multi-metric framework provides a robust methodology for future system validation. Future research should explore disease-specific camera setups and multi-view fusion to leverage the complementary strengths of both perspectives for a more comprehensive gait assessment.

\bibliographystyle{IEEEtran}
\bibliography{ref}

\end{document}